\documentclass[runningheads]{llncs}
\usepackage[T1]{fontenc}

\usepackage{graphicx}
\usepackage{comment}
\usepackage{verbatim}
\usepackage{amsmath}
\usepackage{amssymb}
\usepackage{todonotes}
\usepackage{multirow}
\usepackage[hidelinks]{hyperref}

\usepackage{microtype}
\usepackage{lipsum}

\newcommand{\Object}{Ob}
\newcommand{\Ont}{\mathcal{O}}
\newcommand{\KB}{\mathcal{KB}}
\newcommand{\Concept}{\mathcal{C}}
\newcommand{\Property}{\mathcal{P}}
\newcommand{\Hierarchy}{\mathcal{H}^\mathcal{C}}
\newcommand{\Instance}{\mathcal{I}}
\newcommand{\Lit}{L}
\newcommand{\Annotation}{\mathcal{A}}
\newcommand{\pfn}{prop}
\newcommand{\CInst}{inst^{\Concept}}
\newcommand{\PInst}{inst^{\Property}}

\newcommand{\PatternCoverage}{PC}
\newcommand{\ClassRichness}{CR}
\newcommand{\AveragePopulaton}{P}
\newcommand{\InheritanceRichness}{IR}
\newcommand{\RelationshipRichness}{\overline{RR}}

\newcommand{\RelationshipDiversity}{RD}
\newcommand{\Connectivity}{Cn}
\newcommand{\AverageConnectivity}{\overline{Cn}}
\newcommand{\SiblingFanOutness}{SF}
\newcommand{\ConceptDepth}{CD}
\newcommand{\AverageDepth}{\overline{D}}

\newcommand{\PatternsUsed}{\mathcal{F}_{\text{used}}}
\newcommand{\PatternsTotal}{\mathcal{F}_{\text{tot}}}

\newcommand{\Observation}{\smallskip\noindent\textbf{Observation}:~}
\newcommand{\Requirement}{\smallskip\noindent\textbf{Requirement}:~}
\newcommand{\Metric}[2]{\smallskip\noindent\textbf{Metric}: \emph{#1} (#2).}

\begin{document}

\title{Task-based Ontology Evaluation for Question Generation}

\author{Samah Alkhuzaey\inst{1}\orcidID{0000-0001-8883-1172} \and
    Floriana Grasso\inst{2}\orcidID{0000-0001-8419-6554} \and
    Terry R. Payne\inst{2}\orcidID{0000-0002-0106-8731} \and
    Valentina Tamma\inst{2}\orcidID{0000-0002-1320-610X}}
    \authorrunning{S. Alkhuzaey et al.}
    \institute{Umm al-Qura University, Saudi Arabia \\
    \email{smkhuzaey@uqu.edu.sa} \\
    \and University of Liverpool, Liverpool, UK, L69 3BX \\
    \email{\{F.Grasso, T.R.Payne, V.Tamma\}@liverpool.ac.uk}}

\maketitle              

\begin{abstract}
Ontology-based question generation is an important application of semantic-aware systems that enables the creation of large question banks for diverse learning environments. The effectiveness of these systems, both
in terms of the calibre and cognitive difficulty of the resulting questions, depends heavily on the quality and modelling approach of the underlying ontologies, making it crucial to assess their fitness for this task. To date, there has been no comprehensive investigation into the specific ontology aspects or characteristics that affect the question generation process. 
Therefore, this paper proposes a set of requirements and task-specific metrics for evaluating the fitness of ontologies for question generation tasks in pedagogical settings.
Using the ROMEO methodology (a structured framework used for identifying task-specific metrics), a set of evaluation metrics have been derived from an expert assessment of questions generated by a question generation model.
To validate the proposed metrics, we apply them to a set of ontologies previously used in question generation to illustrate how the metric scores align with and complement findings reported in earlier studies. The analysis confirms that ontology characteristics significantly impact the effectiveness of question generation, with different ontologies exhibiting varying performance levels. This highlights the importance of assessing ontology quality with respect to Automatic Question Generation (AQG) tasks.
\end{abstract}

\section{Introduction}

Research on the use of semantic models within Automatic Question Generation (AQG) has greatly advanced the field of educational assessment.
Ontologies, with their structured representation of knowledge,
introduce an additional layer of semantic understanding which allows AQG systems to generate questions modelled at a semantic level \cite{benedetto2023survey}, thereby surpassing syntax- and vocabulary-focused approaches that operate at a shallower level \cite{gavspar2023evaluation}. Furthermore, semantic-aware AQG models that exploit ontological elements (such as classes and properties) have demonstrated superior generalisability across different domains and question formats, compared to machine learning-based approaches \cite{alkhuzaey2023text}.
Factors such as the size of the ontology, its hierarchical organisation, domain coverage, and defined semantic relations also play a key role in determining the properties of the output \cite{kurdi2020generation,papasalouros2008automatic,stasaski2017multiple,Wang2019OntologyST}; e.g., the linguistic issues in the questions generated have been found to be linked to how concepts are modelled within the ontology \cite{kurdi2020generation}. Yet, to date, there has been no systematic investigation into the type of ontology features that can enhance or hinder pedagogical question generation.

The choice of ontology is crucial when generating and evaluating questions using AQG systems.  
Although it is common practice within the AQG research community to develop experimental ontologies \cite{al2014ontology,alsubait2014generating,diatta2019bilingual,kusuma2020automatic,ev2017automated,venu2020difficulty,e2016modeling}, this has been criticised for potentially affecting the \emph{generalisability} and \emph{validity} of the frameworks and their question patterns, due to them being purpose-built and not representative of real-world applications \cite{alsubait2012automatic,alsubait2014generating}.
Reusing existing ontologies would address these concerns by ensuring that the frameworks had been tested on widely accepted and standardised knowledge structures, thereby enhancing their applicability and reliability across diverse contexts. However, identifying and selecting suitable ontologies that are fit for 
purpose is complex, as this requires the expertise, intuition, and domain knowledge of an ontology specialist. Similarly, when creating new ontologies, frameworks are needed that can
guide construction and be used to evaluate the final product. 
Approaches such as \emph{Application} or \emph{Task-based evaluation} can assess the suitability and fitness of an ontology for a specific task, by evaluating the extent to which an ontology conforms to the task requirements  \cite{brank2005survey,hartmann2005d1,hlomani2014approaches,lourdusamy2018review,pittet2015exploiting}. 
However, for this to be applied to AQG, the requirements (and evaluation metrics) for question generation must first be specified, before they can be used to evaluate how well an ontology supports them.

This paper focuses on identifying the requirements and ontology evaluation metrics that are specifically designed to assess the suitability of an ontology for AQG. By using the ROMEO methodology \cite{yu2009requirements} (a requirements-oriented framework designed for evaluating ontologies in task-specific contexts), a core set of requirements and associated metrics have been identified and evaluated over a set of ontologies.  This expert-led approach combines theoretical and practical perspectives, by integrating domain ontologies into a question-generation model and deriving the relevant metrics based on their observed performance.  
Section \ref{sec:backgound} provides an overview of AQG together with the challenges involved in evaluating ontologies, and the methodology adopted to determine a set of targeted evaluation requirements for AQG is presented in Section \ref{sec:romeo}.
The resulting requirements, evaluation metrics and evaluations are detailed in Section \ref{sec:requirements}, before validating the metrics across a separate set of ontologies used by other AQG studies, to correlate the scores with known question properties in Section \ref{sec:validation}.  The paper concludes in Section \ref{sec:conclusion}.

\section{Background and Related Work}
\label{sec:backgound}

Ontology-based question generation, whereby a domain ontology is used as the knowledge source to generate questions, typically involves the modelling of generic question types using a set of templates containing specific \emph{Resource Description Framework (RDF)} patterns.  
This generates different types of questions using ontological elements such as \emph{concepts} \cite{alsubait2016ontology,diatta2019bilingual}, various types of \emph{properties} \cite{stasaski2017multiple} and \emph{individuals} \cite{diatta2019bilingual}. 
Several of the most common question formats 
are illustrated in Table \ref{tab:patterns}, with the corresponding RDF pattern requirements that vary both in terms of the number of triple patterns they require and the types of elements they include.\footnote{Input ontologies and the resulting questions for each generation strategy are available at \url{https://github.com/SamahAlkhuzaey/Onto_eval_AQG.git}}

Papasalouros \& Chatzigiannakou \cite{papasalouros2018semantic} classify ontology-based questions based on the type of ontological components targeted by the generation strategies:
\emph{Class-based}, \emph{Property-based} and \emph{Terminology-based} (Table \ref{tab:patterns}).
\emph{Class-based questions} assess the learner's understanding of class-instance (i.e. class membership) relationships, such as identifying the class to which a specific entity belongs \cite{al2011ontoque,alsubait2014generating,diatta2019bilingual}, whereas \emph{Property-based questions} focus on testing the learner's knowledge of an entity's properties and their values, such as the attributes or relationships of a given entity \cite{al2011ontoque,papasalouros2008automatic,raboanary2021generating}. \emph{Terminology-based questions} explore hierarchical relationships of an entity within the ontology, such as subclass and superclass associations \cite{alsubait2014generating,cubric2011towards,papasalouros2018semantic}.
Finally, \emph{Annotation-based questions} \cite{alsubait2014generating,raboanary2021generating} leverage annotations in the ontology to generate questions about descriptions or the definitions of terms. These generation strategies result in questions that have a one-to-one mapping between the number of relevant triples and questions generated, and as such, each pattern template only tests the learners' knowledge about a single fact.

Beyond these foundational types, other common question types incorporate additional triples or target special types of semantic relationships, such as \emph{Multiple Choice Questions (MCQs)} and \emph{multi-entity questions}. \emph{MCQs} require not only the generation of a question but also the creation of a set of distractors that are plausible yet distinct from the correct answer \cite{al2014ontology,alsubait2014generating,diatta2019bilingual,leo2019ontology,stasaski2017multiple,vinu2015novel}. \emph{Multi-entity questions} synthesise knowledge from multiple interconnected entities, requiring the learner to have a deeper understanding of the relationships and dependencies between various ontological elements \cite{leo2019ontology,raboanary2022architecture,stasaski2017multiple}. These advanced question types are particularly valuable for evaluating higher-order thinking and comprehension \cite{cubric2020design,leo2019ontology,stasaski2017multiple}, and whilst \emph{MCQs} and multi-entity questions may overlap with the previous categories in their use of classes, properties, or instances, their defining characteristic is the integration of multiple facts to form complex questions and the construction of distractors, that introduces an additional cognitive challenge.

\subsection{Ontology-based Question Generation}
Several studies have explored ontology-based question generation from diverse perspectives by applying these strategies to generate educationally relevant questions. Alsubait et al. \cite{alsubait2012automatic} proposed a structural method for generating analogy-based MCQs. Their system constructs questions in the form \emph{``A is to B as C is to D''} by identifying consistent class–subclass paths in ontologies and scoring the structural similarity of candidate pairs. Distractors are selected based on their structural plausibility whilst maintaining a lower analogy strength than that of the correct answer.

\begin{table}[t]
\caption{Ontology-based question generation patterns and approaches. Uppercase characters represent classes, whereas lowercase ones are instances. $P$ refers to properties.}\label{tab:patterns}
\centering
\setlength{\tabcolsep}{8pt} 
\begin{tabular}{l|l|c}
\textbf{Generation Strategy~} &  \textbf{RDF Pattern} & \textbf{Approaches}\\ 
\hline
\hline

{\bf Class membership} & \texttt{<x> <rdf:type> <X>}  & \cite{al2011ontoque,alsubait2014generating,diatta2019bilingual} \\
\hline

{\bf Property-based} &  \texttt{<x> <P> <y>} & \cite{al2011ontoque,cubric2011towards,papasalouros2008automatic,raboanary2021generating}\\
\hline

{\bf Terminology-based} & \texttt{<x> <rdfs:subClassOf> <y>} & \cite{alsubait2014generating,cubric2011towards,papasalouros2018semantic} \\
\hline

{\bf Annotation-based} & \texttt{<X> <rdfs:comment> <string>} & \cite{alsubait2014generating,raboanary2021generating} \\
\hline

\multirow{3}{*}{\bf MCQs} & \texttt{<x> <rdf:type> <X>} & \multirow{3}{*}{\cite{al2011ontoque,alsubait2014generating,diatta2019bilingual,e2016modeling}} \\  
 & \texttt{<y> <rdf:type> <Y>} & \\  
 & \texttt{<X>, <Y> <rdfs:subClassOf> <Z>} & \\
\hline

\multirow{2}{*}{\bf Multi-entity} & \texttt{<x> <P1> <y> } 
& \multirow{2}{*}{\cite{leo2019ontology,raboanary2022architecture,stasaski2017multiple}} \\  
 & \texttt{<x> <P2> <z>} & 
 
\end{tabular}
\end{table}

Vinu and Kumar \cite{ev2017automated} introduced a template-driven approach for question generation using ontologies, producing class- and property-based questions based on predefined patterns applied across multiple domain ontologies. Their results demonstrated that
the approach was effective in generating a large volume of questions that were semantically comparable to those authored by humans.
Cubric and Tosic \cite{cubric2020design} focused on aligning ontology-based questions with Bloom’s taxonomy \cite{bloom1956taxonomy}, by using ontological structures (particularly class, property, and annotation definitions) to generate questions targeting different cognitive levels, including recall, comprehension, and application. In their evaluation, 69\% of the generated questions were judged suitable for use in assessments, and 33\% were classified as medium to high in cognitive difficulty. The study also reported notable variation in question quality depending on the ontology used, the distractor generation strategy, and the Bloom level targeted; questions that assessed application of knowledge and those based on semantic strategies were rated as highest in quality.
More recently, Raboanary et al. \cite{raboanary2021generating} proposed a flexible architecture for generating a variety of question types, including yes/no, true/false, subclass, equivalence, and definition questions. Their method relies on axiom-based templates and supports both foundational ontology alignment and natural language realisation. The system was evaluated across multiple ontologies and demonstrated the feasibility of producing a wide range of linguistically fluent and structurally valid questions.

\subsection{Ontology Evaluation}
Domain ontologies, which represent the knowledge of a specific subject area, form the primary input for the different question generation models discussed above. Many studies rely on hand-crafted ontologies developed specifically for evaluation purposes \cite{al2014ontology,alsubait2014generating,diatta2019bilingual,kusuma2020automatic,ev2017automated,venu2020difficulty,e2016modeling}, which frequently focus on the knowledge of a particular course  \cite{alsubait2014generating,kusuma2020automatic}, or represent broader knowledge of a general domain \cite{al2014ontology,diatta2019bilingual,kusuma2020automatic,ev2017automated,venu2020difficulty,e2016modeling}. Consequently, such ontologies tend to be small in scope and shallow in structure, capturing only the most general aspects of the domain. This limited depth and scale may constrain the capabilities of question generation models and hinder a comprehensive assessment of their effectiveness. Additionally, the process of building domain-specific ontologies from scratch is time-intensive and requires specialised expertise, posing a challenge to the acceleration of research in this area.
Furthermore, whether a new ontology is engineered or an existing one is reused, it is still important to quantitatively assess if it is fit for the task of question generation through the use of appropriate ontology evaluation metrics.  

Ontology evaluation is a long-standing research area that has evolved together with the growth of ontology engineering and semantic web technologies. Early studies focused on formal evaluation methods that consider aspects such as logical consistency \cite{gangemi2006modelling}, and more recently, broader and more practical considerations are considered such as domain relevance, user-centric adaptability and performance within specific applications \cite{pittet2015exploiting,raad2015survey}. 

Ontologies, as complex knowledge artefacts, can be evaluated across multiple layers to manage their complexity \cite{brank2005survey}: 

\begin{itemize}
    \item \emph{lexical/vocabulary}: examines the modelled concepts and their representation;
    \item \emph{hierarchy/taxonomy}: focuses on subsumption relationships;
    \item \emph{semantic}: checks non-taxonomic relations;
    \item \emph{application}: assesses the ontology's performance and fitness for specific tasks;
    \item \emph{syntactic}: verifies its adherence to the representation language's rules;
    \item \emph{design}: ensures compliance with predefined design principles.
\end{itemize}

As ontologies are used across diverse application contexts, it is crucial that robust evaluation methods are used to assess them before they are adopted
within new systems. Ontology evaluation involves a comprehensive examination of their structure, alignment with domain requirements, and use for specific tasks \cite{hartmann2005d1,hlomani2014approaches}.
However, when reusing an ontology for a specific task, it is imperative to evaluate its suitability and alignment with the task's requirements. 
This is particularly important as ontology evaluation depends heavily on the requirements of the task in question. 

A \emph{requirement} is defined as \emph{``an expression of desired behaviour''} \cite{Pfleeger2005}, with \emph{ontology requirements} specifically denoting the desired quality or competency of an ontology within the context of an application \cite{yu2009requirements}. Given the diversity of ontology requirements across applications, a thorough analysis of the target application needs is essential before ontology reuse. Once these requirements are identified, they can be operationalised \emph{qualitatively}, through expert judgments or user feedback, or \emph{quantitatively}, employing measurable metrics (i.e. measures). Although various evaluation measures for ontologies have been proposed, selecting appropriate measures should be guided by the application's requirements rather than relying on an arbitrary set of metrics \cite{gangemi2006modelling,lozano2004ontometric,park2011ontology,yu2009requirements}.

\section{Requirement-Oriented Methodology for AQG}
\label{sec:romeo}

\begin{table}[t]
\caption{Basic statistics of the ontologies utilised in the expert-led study.}\label{ontos}
\centering
\setlength{\tabcolsep}{0.9pt} 
\begin{tabular}{l|c|c|c|c|c|c}
\multirow{2}{*}{\bf{Ontology}} & 
    \multirow{2}{*}{\bf{Axioms}} &
    \multirow{2}{*}{\bf{Classes}} &
    \multicolumn{3}{| c |}{\bf Number of Properties} &
    \multirow{2}{*}{\bf{Individuals}}\\
&&&
    \bf{Object} &
    \bf{Datatype} & 
    \bf{Annotation} \\  [1ex] 
\hline
\hline
        
\emph{Solar System} &  328 & 10 & 3 & 7 & 1 & 70\\ \hline
\emph{Geography} &  3573 & 9 & 17 & 11 & 3 & 713\\ \hline
\emph{African Wildlife} & 108 & 31 & 5 & 0 & 1 & 0 \\
\end{tabular}
\end{table}

To determine an appropriate set of ontology-evaluation criteria for AQG, it is necessary to first identify those ontology characteristics that are both relevant to the generation of good quality questions, as well as being measurable. \emph{ROMEO (Requirements-Oriented Methodology for Evaluating Ontologies)} is a requirement-oriented approach for identifying a set of unique requirements and characteristics for a task domain (in this context, AQG), that can then be aligned to a set of relevant evaluation metrics \cite{yu2009requirements}.
It was originally adapted from the \emph{Goal-Question-Metrics} (\emph{GQM}) framework \cite{caldiera1994goal}, designed to derive quality measures by linking them to the goals of a given software application.
By determining the requirements of the AQG process through a user-based study, suitable metrics can be identified that quantify the degree to which an ontology can effectively fulfil its intended purpose; i.e. facilitating the generation of high calibre and cognitively challenging questions. 

An expert-led study was therefore conducted to determine a number of ontology requirements given a set of generated questions, based on the following stages:

\begin{enumerate}
    \item 
      A small number of existing domain ontologies were selected for the purpose of  generating questions (Table \ref{ontos}); namely, an astronomy-based ontology focused on the solar system, an ontology representing U.S. geographical data,\footnote{\url{http://www.cs.utexas.edu/users/ml/geo.html}} 
      and an ontology capturing intuitive knowledge about African wildlife \cite{keet2019african}. 
    \item
      Several question generation patterns were utilised to generate questions by instantiating templates with ontology elements (Table \ref{tab:patterns}). This resulted in the generation of 3858 questions\footnote{For MCQs, only unique question stems were considered; repeated questions with varying answer choices were excluded to avoid redundancy.} across all three ontologies (Table \ref{tab:questions}).
    \item 
      An expert familiar with ontology-based AQG was asked to assess the quality of questions, by completing a questionnaire containing a mix of open and closed questions. Two criteria were used to assess the questions: \emph{effectiveness} and \emph{appropriateness} (see below).  Furthermore, they were asked to identify any ontology-related requirements that may impact the quality of the generation process.
    \item 
      The expert responses were qualitatively analysed to gain a comprehensive understanding of the expert's insights, resulting in a set of requirements (discussed in Section \ref{sec:requirements}) that were then used to define the relevant metrics.  
\end{enumerate}

\begin{table}[t]
  \caption{Number of questions generated for each question generation strategy for each ontology in the expert-led study.}
  \label{tab:questions}
  \centering
  \setlength{\tabcolsep}{10pt} 

  \begin{tabular}{l|c|c|c}
    \textbf{} & \emph{Solar System} & \emph{Geography} & \emph{African Wildlife} \\ [1ex]
    
    \hline
    \hline    
    \textbf{Class membership} & 70 & 713 & 0    \\ \hline
    \textbf{Property-based} & 120 & 2044 & 0  \\ \hline
    \textbf{Terminology-based} & 4 & 1 & 25    \\ \hline
    \textbf{Annotation-based} & 10 & 0 & 11   \\ \hline
    \textbf{MCQs} & 8 & 0 & 0   \\ \hline
    \textbf{Multi-entity} & 17 & 724 & 0   \\ 
    \hline
    \hline
    \textbf{Total} & \textbf{351} & \textbf{3471} & \textbf{36}   \\
    
\end{tabular}
\end{table}

The ontologies listed in Table \ref{ontos} were chosen as they exhibit different characteristics and thus could be used to examine how their structural differences influence the question generation outcome, such as an emphasis on taxonomic modelling or hierarchical classifications and conceptual relationships. Two of these were populated to provide rich factual and instance-based representations.
The questions that were presented to the expert were generated using an AQG
approach that utilised the RDF patterns in Table \ref{tab:patterns} within a fixed textual template.  Additional post-processing techniques, such as verbalisation (which are typically carried out to enhance the generation results), were intentionally excluded to maintain consistency in the structure and format of the questions across different ontologies, and to ensure that the expert-led comparison of question quality and effectiveness was not influenced by such techniques.\footnote{Example questions in this paper have been paraphrased for clarity.} 
The expert reviewed the question specifications (Table \ref{tab:patterns})
and evaluated the questions generated from each of the three ontologies, with the aim of identifying specific ontological features that contributed to variations in the questions. The two evaluation criteria employed by Leo et al. \cite{leo2019ontology} were used when assessing each ontology and its corresponding questions:
\begin{itemize}
    \item
        The \emph{effectiveness} of the questions is assessed by considering: 1) the number and variety of questions generated; 2) addressing whether the ontology produced all question types; and 3) comparing the quantity of questions across ontologies and templates. This is done by analysing the questions in relation to defined ontological relationships. 
    \item
        The \emph{question quality (appropriateness)} is assessed based on how pedagogically appropriate the questions are, to assess learners at different knowledge levels by ensuring questions range 
        from basic to a more advanced understanding of the domain.
\end{itemize}

\begin{table*}[t]
\caption{Symbols used by the metrics following the nomenclature used by Tartir et.al. \cite{tartir2005ontoQA,tartir2010ontological}.}\label{tab:symbols}
\centering
\begin{tabular}[t]{lp{0.43\linewidth}}
    \multicolumn{2}{c}{\bf Ontology Schema} \\
    \multicolumn{2}{c}{$\Ont = \{\Concept, \Property, \Hierarchy, \pfn, \Annotation\}$} \\ [1ex]

\hline
\hline
    $\Concept$ & a set of \emph{concepts}, disjoint from $\Property$.   $\Concept' \subseteq \Concept $ denotes the subset of \emph{populated concepts} (i.e., concepts that have at least one instance $\Instance$ in $\KB$). \\

    $\Property$ & a set of \emph{properties}, disjoint from $\Concept$. \\

    $\Hierarchy$ & a \emph{concept hierarchy}, which is a directed, transitive relation, and relates to the taxonomy. $\Hierarchy(c_1, c_2)$ states that $c_1$ is a sub-concept of $c_2$ (i.e. $c_1 \subseteq c_2$). \\

    $\pfn$: & $\Property \to \Concept \times \Concept$: A function mapping properties that link pairs of concepts. $prop(p) = (c_1,c_2)$ can also be expressed as $p(c_1, c_2)$. \\

    $\Annotation$ & a set of annotation properties of the type \texttt{rdfs:comment}. \\

\end{tabular}
~~~
\begin{tabular}[t]{lp{0.36\linewidth}}
    \multicolumn{2}{c}{\bf Knowledge Base} \\
    \multicolumn{2}{c}{$\KB = \{\Ont, \Instance, \Lit, \CInst, \PInst\}$} \\ [1ex]

\hline
\hline
    $\Ont$ & the Ontology Schema. \\
    $\Instance$ & a set of instance identifiers disjoint from $\Concept$ and $\Property$. \\

    $\Lit$ & a set of literal values. \\
        
    $\CInst$: & $\Concept \to 2^{\Instance}$ (concept instantiation) maps concepts to their instances. $\CInst(c) = i$ may also be written as $c(i)$. \\
        
    $\PInst$: & $\Property \to 2^{\Instance \times \Object}$.
    For each property $p \in  \Property$,  $\PInst$ is the set of pairs $(i,o)$ such that the assertion $p(i,o)$ holds, with $i \in \Instance$ and $o \in \Object = \Instance \cup \Lit$. This may also be written as $p(i,o)$.
    
\end{tabular}
\end{table*}

As this evaluation task involves an extensive qualitative analysis of a large pool of questions generated from three diverse ontologies, a single expert was recruited by this study to assess the feasibility of the proposed approach, as well as to understand critical aspects such as the time and effort needed by experts for such evaluation. 
The recruitment of a singe expert was sufficient in this case as the objective was to identify structural and semantic features of ontologies through pedagogically relevant questions, rather than to establish consensus or benchmark performance.

\section{Response Analysis and Requirement Derivation} \label{sec:requirements}

This section presents the outcomes of the expert-led study, by describing the {\bf observations} made by the expert, the {\bf requirements} that the expert derived from these observations, and proposed {\bf metrics} that were subsequently defined to measure these requirements. The proposed evaluation metrics build upon well-established measures \cite{gangemi2006modelling,Maedche2002,tartir2005ontoQA,tartir2010ontological} that have been adapted to align with the specific requirements of the AQG task.
The model used for defining metrics formally specifies the ontology schema and the knowledge base structures.
An ontology schema represents the class and property definitions (i.e. {\bf T-box}) and is denoted by the 5-tuple $\Ont = \{\Concept, \Property, \Hierarchy, \pfn, \Annotation\}$, whereas a knowledge base (i.e. {\bf T-box $\cup$ A-Box}) represents the instantiated ontological elements (individuals) and is denoted by the 5-tuple $\KB = \{\Ont, \Instance, \Lit, \CInst, \PInst\}$. The meaning of the symbols used in these definitions are given in Table \ref{tab:symbols}. Whilst this model is similar to others proposed for various ontology scenarios \cite{Maedche2002}, it has been adapted to be more pertinent to the metrics that focus on AQG tasks.

\subsection{RQ1: Include essential RDF pattern pre-requisites.}

\Observation
The first analysis examined whether the ontologies could generate all of the question types listed in Table \ref{tab:patterns}. The \emph{Geography} ontology was the largest considered (in terms of axioms), and consequently generated the highest number of questions (Table \ref{tab:questions}), although the majority of these were property-based. However, size alone does not determine its suitability for the task; whilst obviously influential, structural complexity is also equally (if not more critically) important in accommodating the diverse axiom requirements of different question types.
For example, no annotation-based questions were generated for \emph{Geography} as it lacked annotation properties that could match the pattern \texttt{<x> <rdfs:comment> <string>}. 
The \emph{Solar System} ontology generated questions for each of the types, reflecting a comprehensive structure and completeness in satisfying the question pattern pre-requisites from Table \ref{tab:patterns}. This contrasts with \emph{African Wildlife}, which suffered from a lack of individuals, and only generated terminology-based and annotation-based questions. 
These observations emphasise that structural complexity, including the presence of all prerequisites (e.g. the patterns in Table \ref{tab:patterns}) are essential for generating a comprehensive range of question types. 
Missing axioms or incomplete structures can significantly limit the ontology's ability to support specific templates, hindering its effectiveness for question generation.

\Requirement
An input ontology should assist in generating all desirable question types, regardless of the axiom prerequisites required for each question. 
This requirement can be evaluated through several metrics that address two key questions:
\begin{enumerate}
    \item \emph{``How many pattern fragments align with the components of the ontology schema and metadata model?''}
    \item \emph{``What is the structural complexity level of each component in the ontology?''}
\end{enumerate}

\Metric{Pattern Coverage}{$\PatternCoverage \in [0,1]$}
The first question assesses the completeness of the ontology based on its utilisation of RDF pattern fragments (Table \ref{tab:patterns}), described by the instantiation of concepts ($\CInst$) and objects ($\PInst$), the class hierarchy ($\Hierarchy$) and schema annotations ($\Annotation$). The metric, defined as
$\PatternCoverage = \PatternsUsed / \PatternsTotal$ quantifies the ratio of pattern fragments used ($\PatternsUsed$) to the total available defined within the AQG patterns ($\PatternsTotal$), indicating how effectively the ontology leverages its formal language. A high score suggests full pattern utilisation, while a lower score identifies missing elements that may hinder question generation (see Table \ref{tab:metrics}, together with the scores for each ontology evaluated).

\emph{Solar System} achieved full coverage ($\PatternCoverage = 1$), aligning with its ability to generate all question types (Table~\ref{tab:questions}). \emph{Geography} scored lower due to missing pattern types, particularly the absence of annotation properties (\texttt{rdfs:comment}), leading to $\Annotation = \varnothing$. \emph{African Wildlife} covered only half of the patterns, limiting its question generation to taxonomic and annotation types.

\Metric{Class Richness}{$\ClassRichness \in [0,1]$} The second question (\emph{``What is the structural complexity level of each component in the ontology?''}) extends the first by assessing \emph{structural complexity} at both schema and data levels through two key metrics:
Class Richness and Average Population (adopted from Tartir et al. \cite{tartir2005ontoQA,tartir2010ontological}). \emph{Class Richness} ($\ClassRichness = |\Concept'| / |\Concept|$) measures the ratio of classes in the ontology ($\Concept$) to those that have been instantiated within the Knowledge Base ($\Concept'$); which is essential for generating class-based questions (Table \ref{tab:metrics}). A low $\ClassRichness$ indicates insufficient instantiation of classes within the KB, limiting its suitability for question generation.

Ontologies with well-populated KBs, such as \emph{Solar System} and \emph{Geography}, achieved high $\ClassRichness$, while \emph{African Wildlife}, lacking instances, scored zero.

\Metric{Average Population}{$\AveragePopulaton \in \mathbb{R}$} This metric measures the average number of instances per class, indicating whether classes are sufficiently populated ($P = |\Instance| / |\Concept|$). This directly affects class-based question generation and other instance-dependent templates (e.g. property-based strategies).
Together with $\ClassRichness$, it reveals the ontology population distribution, influencing both question diversity and volume. 

For example, \emph{Geography} (79 instances per class) generated 713 questions, yet \emph{Solar System} (7 instances per class) produced only 70.

\Metric{Inheritance Richness}{$\InheritanceRichness \in \mathbb{R}$}
As terminology-based questions are solely dependent on the taxonomic relationships defined within the ontology, it is necessary to quantify the ratio between the total number of subclasses ($|\Concept_1|$) and the total number of classes $|\Concept|$ defined in the ontology schema. This metric determines the average number of subclasses per class
($\InheritanceRichness = {|\Concept|}^{-1}  \sum_{c \in \Concept} \left| \Hierarchy (c_1, c_2) \right|$).

\emph{Solar System} achieved the highest score ($\InheritanceRichness = 0.9$), indicating a denser use of subclassing despite its smaller size, while \emph{African Wildlife} scored slightly lower ($\InheritanceRichness = 0.8$) due to a larger number of top-level classes.

\Metric{Relationship Diversity}{$\RelationshipDiversity \in [0,1]$}
To evaluate the ontology’s capability to generate property-based questions, two additional metrics are examined: \emph{Relationship Diversity} and \emph{Relationship Richness}. This question generation strategy relies on the use of properties $\Property$ to connect objects
$\Object = \Instance \cup \Lit$ (i.e. instances $\Instance$ or literals $\Lit$) thus targeting the instantiated properties $\PInst$ in the knowledge base, resulting in the formation of more detailed questions regarding concepts' properties.
\emph{Relationship Diversity} reflects the balance between taxonomic and non-taxonomic relations, and is defined as
$\RelationshipDiversity = |\PInst| ~/~ (|\Hierarchy|+|\PInst|)$. A high $\RelationshipDiversity$ indicates an emphasis on non-taxonomic relations, increasing property-based questions while reducing terminology-based ones.

\emph{Solar System} and \emph{Geography} scored highly, favouring property-driven questions. However, \emph{African Wildlife} ($\RelationshipDiversity = 0$) prioritised taxonomic knowledge but, as it is lacking in instances, generated no questions of either type (Table~\ref{tab:questions}).

\Metric{Average Relationship Richness}{$\RelationshipRichness \in \mathbb{R}$}
This quantifies the number of properties defined in the schema for each class and that are utilised by instances in the KB, by first computing a relationship richness value for each class ($RR_{c_i}$), and then taking the mean across all classes. This is defined as the ratio of the number of properties used by the instances of class $c_i$ (i.e. $|p(i,o)|$) and the total number of properties defined for class $c_i$ in the schema (i.e. $|p(c_i,c_j)|$):
\[
\RelationshipRichness = \frac{1}{|\Concept|} \sum_{c_i \in \Concept} RR_\Concept, \text{~~~~where~}
RR_{c_i} = \frac{| p(i, o)|, i \in c_i(\Instance)}{\left| p(c_i,c_j)\right| }
\]
An ontology with a high $\RelationshipRichness$ (e.g. $\RelationshipRichness = 0.5$ and $\RelationshipRichness = 0.9$ for \emph{Solar System} and \emph{Geography} respectively) indicates a well-structured 
axiomatic foundation aligned with the prerequisites, highly suited for generating property-based questions.

\subsection{RQ2: Supports the generation of multi-entity questions.}

\Observation
The \emph{Geography} ontology generated the largest number of multi-entity questions (724 questions); e.g. \emph{``What is the highest point in Georgia with an elevation of 1,458 meters?''}. The expert attributed this to the rich and diverse range of descriptions within that ontology relating to the breadth of properties used to model each concept; for example, the concept \texttt{State} in \emph{Geography} included the properties \texttt{statePopulation}, \texttt{stateArea} and \texttt{borders}.
Likewise, \emph{Solar System}, though much smaller in size, generated 17 multi-entity questions, again due to having a broad set of properties. However, no multi-entity questions were generated for \emph{African Wildlife}, as it lacked instances in its KB.

\Requirement
The effectiveness of an ontology should consider support for the generation of the more advanced multi-entity questions.  This type of question involves the use of more than one axiom, allowing for greater variation in question generation, thus testing learners' knowledge of how concepts interact \cite{stasaski2017multiple}. This variation is also important as it enables the creation of questions that go beyond simple, one-fact questions, offering a broader and deeper understanding of the domain (i.e. targeting the second level in Bloom's taxonomy \cite{bloom1956taxonomy,stasaski2017multiple}). Multi-entity questions tend to be more detailed and complex, and cover a wider breadth of knowledge within the domain. These questions are generated when an ontology contains rich entity descriptions, with multiple axioms that comprehensively define an entity by detailing its properties and associations \cite{stasaski2017multiple}.

\Metric{Average Connectivity}{$\AverageConnectivity \in \mathbb{R}$}
The fact that multi-entity questions are formed by combining multiple facts regarding an entity poses the question \emph{``How well are instances connected to other objects?''} The corresponding metric should therefore capture the notion of \emph{Connectivity} ($\Connectivity$), which quantifies the \emph{out-degree} of an instance (i.e the number of property assertions it makes). This reflects how richly described instances are within the ontology, and is defined as:

\[
\AverageConnectivity = \frac{1}{|\Instance|} \sum_{i \in I} |p(i, o)|
\]

\noindent
where $\Instance$ is the set of all instances in the ontology, and $p(i, o)$ denotes the set of outgoing property assertions (both object and datatype) made by instance $\Instance_i$. Thus, \emph{Average Connectivity} captures the overall degree of factual richness at the instance level.

\emph{Geography} scores highly ($\AverageConnectivity = 2.8$) 
and \emph{Solar System} less so but with a positive score ($\AverageConnectivity =  1.7$), suggesting that both are well-suited for generating multi-entity questions, which typically involve 2–3 connections per question.


\begin{table}[t]
\caption{Comparative evaluation of ontologies across the proposed evaluation metrics.}\label{tab:metrics}
\centering 
  \setlength{\tabcolsep}{3pt} 
\renewcommand{\arraystretch}{1.4}

\begin{tabular}{ll|c|c|c}

\multirow{2}{*}{\bf{Evaluation Name}} & \multirow{2}{*}{\bf Metric} & \emph{Solar} & \multirow{2}{*}{\emph{Geography}} & \emph{African} \\

 & &  \emph{System} & & \emph{Wildlife}\\ [1ex]
\hline
\hline
 Pattern Coverage & 
 $\PatternCoverage = \tfrac{\PatternsUsed}{\PatternsTotal}$
 & 1 & 0.7 & 0.5 \\

 Class Richness &
 $\ClassRichness = \tfrac{|\Concept'|}{|\Concept|}$ & 0.8 & 1 & 0 \\ 

 Average Population &
 $P = \tfrac{|\Instance|}{|\Concept|}$
 & 7 & 79 & 0 \\ 

Inheritance Richness &
$\InheritanceRichness = \tfrac{1}{|\Concept|} \sum_{c \in \Concept} \left| \Hierarchy (c_1, c_2) \right|$
& 0.9 & 0.1 & 0.8 \\ 

Relationship Diversity &
$\RelationshipDiversity = \tfrac{|\PInst|}{|\Hierarchy|+|\PInst|}$
& 0.9 & 0.9 & 0 \\ 

Av. Relationship Richness &
$\RelationshipRichness = \tfrac{1}{|\Concept|} \sum_{\Concept_i \in \Concept} RR_C$
& 0.5 & 0.9 & 0 \\ 

Average Connectivity &
$\AverageConnectivity = \frac{1}{|\Instance|} \sum_{i \in I} |p(i, o)|)$
& 1.7 & 2.8 & 0 \\  

Sibling fan-outness &
$    SF = \frac{|C'_{\text{sib}}|}{|C|}$
& 0.8  & 0  & 0 \\

Average Depth &
$\AverageDepth = \frac{1}{|\Concept|} \sum_{\Concept_i \in \Concept} \ConceptDepth(\Concept_i)$
& 2.3 & 1.1 & 1.9 \\


 \end{tabular}
 \vspace{-0.1in}
\end{table}


\subsection{RQ3: Supports the generation of plausible MCQ distractors.} \label{subsec:distractors}

\Observation
Generating MCQs posed challenges across all three ontologies. Questions of the form \emph{``Which of these is X?''} typically list the answer choices with the correct answer as a subclass of $X$ and incorrect answers as sibling classes of $X$. No MCQs were generated for \emph{African Wildlife}, due to a lack of instances within the sibling classes, whereas 
the reason for the lack of MCQs for \emph{Geography} was that no sibling classes were identified.
\emph{Solar System} contains a rich underlying knowledge graph with each class having several sub-classes, and thus well-defined sibling relationships. Of the three ontologies, this was the only one that successfully generated MCQs.

\Requirement
MCQs are the most common question types generated as they offer a more complex structure that requires the incorporation of several axioms with special characteristics. The structure of an ontology plays a key role in this process, particularly in the generation of plausible distractors for MCQs. 
These distractors are crafted to resemble the correct answer by leveraging semantic relationships within the ontology. Specifically, sibling classes (i.e. those that share a common immediate parent) are ideal candidates for distractors due to their inherent semantic similarity \cite{diatta2019bilingual,papasalouros2008automatic}.
In order to generate plausible distractors, an ontology should contain a hierarchical structure where populated sibling classes are well-represented, leading to the question \emph{``Does the ontology contain populated classes with a sufficient number of sibling classes?''}. 

\Metric{Sibling Fan-Outness}{$\SiblingFanOutness \in \mathbb{R}$}
 This is measured by calculating the ratio of the absolute populated sibling cardinality ($\Concept'_{\text{sib}}$) to the total number of populated classes ($\Concept'$) in the ontology; i.e.
$SF = |C'_{\text{sib}}| / |\Concept'|$.
A high $\SiblingFanOutness$ score indicates that the ontology has a horizontally branching structure with multiple sibling classes under a few superclasses.
This results in a broad hierarchy where categories are general rather highly differentiated. Such a structure is particularly effective for generating MCQ distractors with close semantic similarity.

\emph{Solar System} was the only ontology to exhibit this pattern and successfully generated MCQs ($\SiblingFanOutness=0.8$), whereas as \emph{Geography} lacked classes that shared a common parent, no MCQs were generated.  Likewise, as \emph{African Wildlife} lacked any instances in its KB, it also generated no MCQs ($\SiblingFanOutness=0$).

\subsection{RQ4: Supports the generation of questions with varied conceptual diversity.} 

\Observation
An analysis of question quality with a focus on cognitive challenge considered how questions varied with respect to the level of knowledge required to successfully answer the question as evaluated by the expert. While several factors affect cognitive complexity, the emphasis here is on ontology-driven factors rather than external complexities such as the domain itself. 
When examining the types of questions generated from taxonomic relationships, the expert noted that \emph{Geography} generated questions about high-level domain concepts (e.g. \emph{City} and \emph{River}). In contrast, \emph{African Wildlife} and \emph{Solar System} generated questions targeting both broad and more specific concepts.
Kurdi \cite{kurdi2020generation} demonstrated that the diversity of generated questions depends not only on the size of the ontology but also on the richness of its hierarchical structure.

\Requirement
A well-structured ontology should support conceptual diversity by enabling question generation across different levels of understanding, from fundamental to complex. This ensures broad domain coverage and facilitates the generation of cognitively diverse questions. Several studies have also considered conceptual depth as an important characteristic that directly impacts the ability to create cognitively diverse questions \cite{faizan2018automatic,faizan2017multiple,kuo2003difficulty,kuo2004analyzing,e2016modeling,venu2020difficulty}.

\Metric{Average Depth}{$\AverageDepth \in \mathbb{R}$}
This requirement can be assessed by
considering whether the ontology includes concepts from varying depths of the hierarchical structure. To evaluate this, $\AverageDepth = {|\Concept|}^{-1} \sum_{c_i \in \Concept} \ConceptDepth(c_i)$ is used to where $\ConceptDepth(c_i)$ is the depth of concept $c_i$ in the hierarchy $H^c$ (defined as the edges from $c_i$ to the root).
This metric (similar to that defined by Gangemi et al. \cite{gangemi2006modelling}) yields a single average depth across all concepts to ensure question generation spans general to specific concepts, highlighting that hierarchical richness, rather than size alone, drives cognitive diversity.

Both \emph{Solar System} ($\AverageDepth = 2.3$) and \emph{African Wildlife} ($\AverageDepth = 1.9$) exhibited rich hierarchies, generating questions at varying conceptual levels. For example, \emph{Solar System} included broad concepts like \emph{``Planet''} while deeper levels introduced distinctions such as \emph{``Terrestrial Planet''} and \emph{``Gas Giant''}.

\section{Empirical Validation of the Proposed Metrics} \label{sec:validation}

\begin{figure}[t]
\centering
\includegraphics[width=\linewidth]{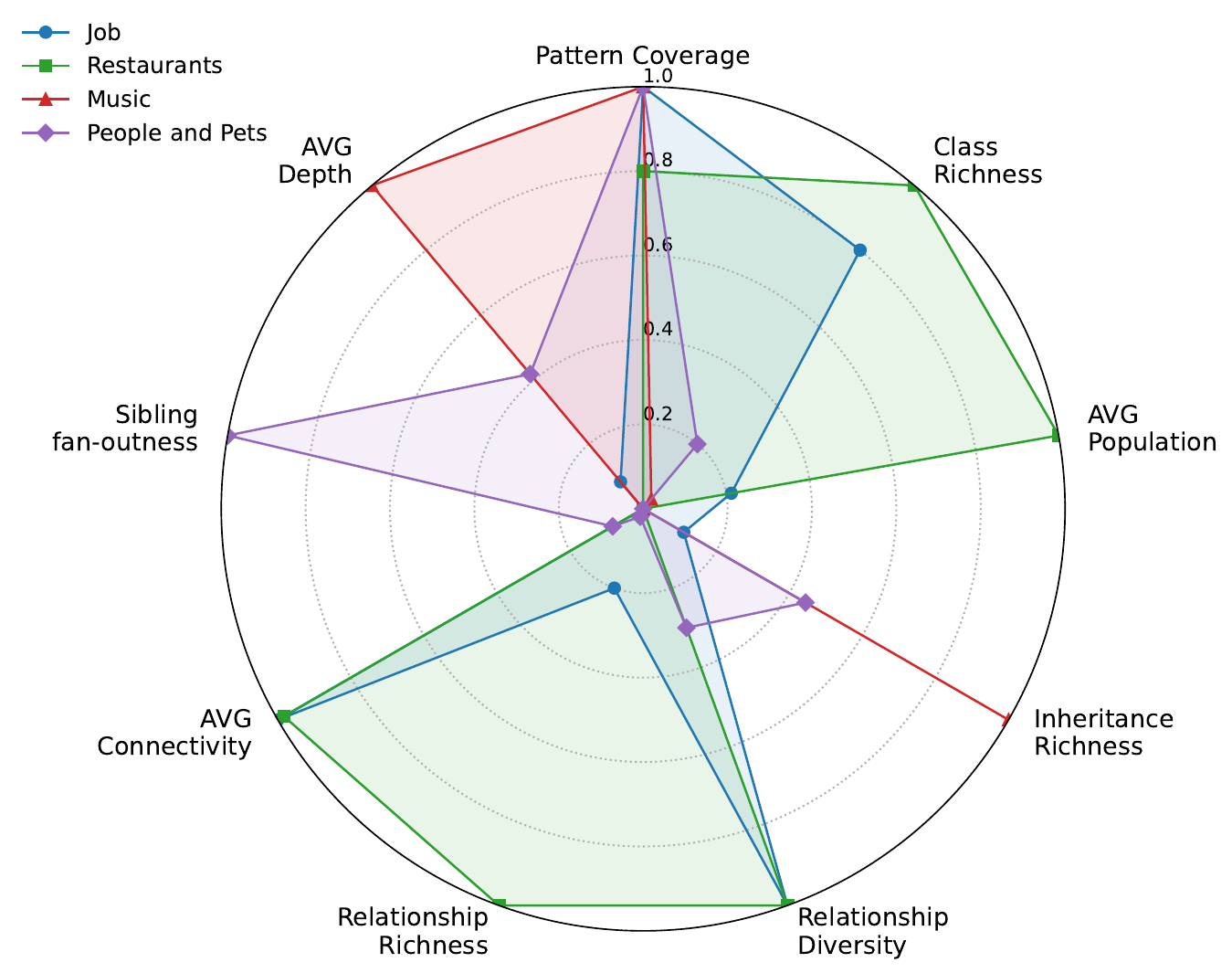}
\caption{Normalised results for the proposed ontology evaluation metrics applied to the \emph{Job}, \emph{Restaurant}, \emph{Music}, and \emph{People \& Pets} ontologies}
\label{fig:onto_chart}
\end{figure}

To validate the metrics within the context of AQG, 
four ontologies that had previously been used in ontology-based question generation studies were selected (\emph{Music}~\cite{cubric2020design}, \emph{Job}~\cite{ev2017automated},  \emph{Restaurant}~\cite{ev2017automated} and \emph{People \& Pets}~\cite{alsubait2012automatic}) and evaluated both in terms of their profile (using the metrics summarised in Table \ref{tab:metrics}), and the reported characteristics of the questions generated by each study. In each case, both the ontology and their associated knowledge bases were analysed, and the values for each metric were calculated (Table \ref{tab:new_metrics}). 
The radar diagram (Figure \ref{fig:onto_chart}) provides a comparative analysis of the normalised profiles (as defined by the different metrics) for each ontology. Higher values indicate more developed features, such as dense class instantiation, diverse relationships, or deeper class hierarchies. By analysing each of these profiles, we demonstrate that the metric scores (and hence the metric definitions themselves) align with and complement the prior findings of other independent studies, thus validating the relevance of the proposed metrics for ontology-based AQG approaches.

Cubric and Tosic \cite{cubric2020design} utilised the \emph{Music} Ontology for question generation and reported notable limitations in producing questions that relied on Property-based and Class-based strategies (particularly MCQs). These generation strategies rely heavily on the availability of richly instantiated data, specifically well-defined class-instance relationships (e.g. \texttt{<ex:TheBeatles> <rdf:type> <mo:MusicArtist>}) and instance-property assertions (e.g. \texttt{<ex:TheBeatles> <mo:genre> <ex:Rock>}). However, such relationships are largely absent from the ontology itself, which primarily encodes schematic structures rather than populated instance data. This under-representation is directly reflected in the different values for the evaluation metrics in Table \ref{tab:new_metrics}.

\emph{Music} resulted in low scores across several key evaluation metrics. Notably, its \emph{Class Richness} is limited, with only 3\% of classes populated with instances. Moreover, the \emph{Average Population} per class is extremely low, at just 0.02 instances per class. Compounding this, the \emph{Average Relationship Richness} metric indicates that none of the properties defined at the schema level are utilised to describe instances at the data level. Collectively, these metrics reflect a low degree of instantiation and limited interconnection among instances within the ontology. This supports Cubric and Tosic's observation that their AQG approach performed poorly on \emph{Music} for Class-based and Property-based question generation, largely due to the absence of meaningful associations between classes and instances \cite{cubric2020design}.

The results demonstrate that \emph{Music} is primarily schematic (as further indicated by its relatively high \emph{Inheritance Richness}, \emph{Relationship Diversity} and \emph{Average Depth} scores), encoding structural knowledge about musical concepts (e.g., \texttt{Genre}, \texttt{Instruments} and \texttt{Lyrics}) rather than detailed instance-level data. Consequently, it is better suited to supporting Terminology-based questions, which depend predominantly on class hierarchies and labels, and is less appropriate for generating Property-based, Class-based or Multi-entity questions that require a richer population of instances. Thus, the metric results align closely with the ontology’s reported performance in prior AQG experiments.

Vinu and Kumar \cite{ev2017automated} utilised the \emph{Job} and \emph{Restaurant} ontologies for Class-based and Property-based question generation. They reported a high number of relevant triples (triples retrieved from the ontologies that satisfy question requirements); for example, over 288,000 triples were retrieved from the \emph{Restaurant} ontology using a single question pattern. These triple counts serve as an indicator of the ontologies’ potential to support large-scale question generation and reflect their extensive instance populations.

This observation is supported by the results of the evaluation metrics in Table \ref{tab:new_metrics}, which show that both ontologies exhibit high \emph{Class Richness} and \emph{Average Population} scores. The high \emph{Class Richness} scores for \emph{Job} ($\ClassRichness = 0.8$), and for \emph{Restaurant} ( $\ClassRichness = 1$) indicate that almost all classes are populated with instances. Both \emph{Restaurant} and \emph{Job} are also sufficiently populated; however, the metrics for 
\emph{Restaurant} suggest a better utilisation of defined properties at the schema level, with a \emph{Average Relationship Richness} of $\RelationshipRichness=0.5$ compared to \emph{Job} ($\RelationshipRichness=0.1$), which indicates a higher level of instance interconnection. 
The structural analysis also reveals that both schemata are relatively flat. In particular, the \emph{Sibling Fan-Outness} metric is low, suggesting that although the classes are well populated, the ontologies lack a broad class hierarchy with a sufficient number of sibling classes. 
This limitation has practical implications. Whilst \emph{Job} and \emph{Restaurant} are well suited to generating question stems for Class-based and Property-based questions, they are less effective for generating MCQs with semantically plausible distractors, a process that typically relies on the presence of sibling classes (Section \ref{subsec:distractors}). Notably, Vinu and Kumar's study \cite{ev2017automated} did not attempt full MCQ generation with distractors from these ontologies, focusing instead on stem generation, which is consistent with the structural characteristics revealed by this metric-based analysis.

The \emph{People \& Pets} ontology was employed by Alsubait et al. \cite{alsubait2012automatic} to generate analogy-style, Terminology-based MCQs that rely on structural patterns such as class–subclass and sibling relationships. As illustrated in Table \ref{tab:new_metrics}, the evaluation metrics align with this intended use, but also explain the relatively limited output (only 15 questions were generated from this ontology) \cite{alsubait2012automatic}.
The \emph{Inheritance Richness} score ($\InheritanceRichness=0.4$) and \emph{Average Depth} ($\AverageDepth=1.5$) suggest that the ontology possesses only a moderate hierarchical structure, which probably constrained the diversity of eligible concept pairs, thus contributing to the small number of generated analogies.

More critically, the ontology exhibits a high \emph{Sibling Fan-Outness} score ($\SiblingFanOutness=0.5$), indicating that several classes are positioned under shared superclasses with sufficient horizontal breadth. This structure is particularly valuable for generating MCQs with semantically plausible distractors, as distractors in analogy questions often draw on sibling concepts to maintain a coherent semantic frame. This structural trait likely played a central role in enabling the generation of MCQs, even if the overall quantity was low.
The ontology’s low \emph{Class Richness} ($\ClassRichness=0.2$) and \emph{Average Population} ($\AveragePopulaton=0.3$) further confirm that it is schema-focused and sparsely populated with instances. Additionally, the \emph{Relationship Diversity} and \emph{Average Relationship Richness} scores ($\RelationshipDiversity=0.3$ and $\RelationshipRichness=0.01$, respectively) are low, reinforcing the notion that the ontology supports limited relational inference beyond subclass structures.
Thus, the \emph{People \& Pets} ontology demonstrates structural properties suitable for Terminology-based MCQ generation. While the relatively shallow hierarchy limited the number of analogies that could be extracted, the strong presence of populated sibling structures made it especially effective for generating MCQs with meaningful distractors \cite{alsubait2012automatic}.

\begin{table}[t]
\caption{Comparative evaluation of the validation ontologies across the proposed evaluation metrics.}\label{tab:new_metrics}
\centering 
\setlength{\tabcolsep}{6pt}
\renewcommand{\arraystretch}{1.1}

\begin{tabular}{l|c|c|c|c}

\bf{Evaluation Metric} & \emph{Music} & \emph{Restaurant} & \emph{Job} & \emph{People \& Pets} \\ [1ex]
\hline
\hline
Pattern Coverage ($\PatternCoverage$) & 1 & 0.8 & 1 & 1 \\
Class Richness ($\ClassRichness$) & 0.03 & 1 & 0.8 & 0.2 \\
Average Population ($\AveragePopulaton$) & 0.2 & 2436 & 517 & 0.3 \\
Inheritance Richness ($\InheritanceRichness$) & 0.9 & 0 & 0.1 & 0.4 \\
Relationship Diversity ($\RelationshipDiversity$) & 0 & 1 & 1 & 0.3 \\
Av. Relationship Richness ($\RelationshipRichness$) & 0 & 0.5 & 0.1 & 0.01 \\
Average Connectivity ($\AverageConnectivity$) & 0 & 5.9 & 6 & 0.5 \\
Sibling Fan-Outness ($\SiblingFanOutness$) & 0 & 0 & 0 & 0.5 \\
Average Depth ($\AverageDepth$) & 2.2 & 1 & 1.1 & 1.5 \\

\end{tabular}
\end{table}

\section{Conclusion}
\label{sec:conclusion}

In this paper, we address the need for a way to characterise ontologies for use in ontology-based Automatic Question Generation. We propose a set of metrics, identified by using the ROMEO methodology that defines metrics based on specific task requirements.  By consulting an expert in pedagogy, we defined a set of requirements and associated metrics to assess how different ontologies perform in AQG tasks.  The metrics were then validated across an independent set of ontologies and their associated knowledge bases used in previous ontology-based AQG studies. Our results demonstrate that the characteristics of ontologies significantly influence the effectiveness of the question generation process, and that these characteristics are reflected accurately by these metrics. Our findings underscore that different ontologies yield varying levels of performance, highlighting the critical need to assess ontology quality in AQG. 

%
%
%
\bibliographystyle{splncs04}
\bibliography{aied2025}

\begin{thebibliography}{10}
\providecommand{\url}[1]{\texttt{#1}}
\providecommand{\urlprefix}{URL }
\providecommand{\doi}[1]{https://doi.org/#1}

\bibitem{al2011ontoque}
Al-Yahya, M.: {OntoQue}: {A Question Generation Engine for Educational Assesment based on Domain Ontologies}. In: the IEEE 11th International Conference on Advanced Learning Technologies. pp. 393--395. IEEE (2011)

\bibitem{al2014ontology}
Al-Yahya, M.: Ontology-based multiple choice question generation. The Scientific World Journal  \textbf{2014}(1) (2014)

\bibitem{alkhuzaey2023text}
AlKhuzaey, S., Grasso, F., Payne, T.R., Tamma, V.: Text-based question difficulty prediction: A systematic review of automatic approaches. International Journal of Artificial Intelligence in Education  \textbf{34},  862--914 (2023)

\bibitem{alsubait2012automatic}
Alsubait, T., Parsia, B., Sattler, U.: Automatic generation of analogy questions for student assessment: an ontology-based approach. Journal of Research in Learning Technology  \textbf{20},  95--101 (2012)

\bibitem{alsubait2014generating}
Alsubait, T., Parsia, B., Sattler, U.: Generating multiple choice questions from ontologies: Lessons learnt. In: the 11th OWL: Experiences and Directions Workshop {(OWLED} 2014). vol.~1265, pp. 73--84 (2014)

\bibitem{alsubait2016ontology}
Alsubait, T., Parsia, B., Sattler, U.: Ontology-based multiple choice question generation. KI-K{\"u}nstliche Intelligenz  \textbf{30}(2),  183--188 (2016)

\bibitem{benedetto2023survey}
Benedetto, L., Cremonesi, P., Caines, A., Buttery, P., Cappelli, A., Giussani, A., Turrin, R.: A survey on recent approaches to question difficulty estimation from text. ACM Computing Surveys  \textbf{55}(9),  1--37 (2023)

\bibitem{bloom1956taxonomy}
Bloom, B.S., Englehart, M.D., Furst, E.J., Hill, W.H., Krathwohl, D.R.: Taxonomy of educational objectives, handbook I: the cognitive domain. New York: David McKay Co Inc (1956)

\bibitem{brank2005survey}
Brank, J., Grobelnik, M., Mladenic, D.: A survey of ontology evaluation techniques. In: the conference on data mining and data warehouses (SiKDD 2005). pp. 166--170. Citeseer (2005)

\bibitem{caldiera1994goal}
Caldiera, V.R.B.G., Rombach, H.D.: The goal question metric approach. Encyclopedia of software engineering pp. 528--532 (1994)

\bibitem{cubric2011towards}
Cubric, M., Tosic, M.: Towards automatic generation of e-assessment using semantic web technologies. International Journal of e-Assessment  \textbf{1}(1) (2011)

\bibitem{cubric2020design}
Cubric, M., Tosic, M.: Design and evaluation of an ontology-based tool for generating multiple-choice questions. Interactive Technology and Smart Education  \textbf{17}(2),  109--131 (2020)

\bibitem{diatta2019bilingual}
Diatta, B., Basse, A., Ouya, S.: Bilingual ontology-based automatic question generation. In: the IEEE Global Engineering Education Conference (EDUCON). pp. 679--684. IEEE (2019)

\bibitem{faizan2018automatic}
Faizan, A., Lohmann, S.: Automatic generation of multiple choice questions from slide content using linked data. In: the 8th International Conference on Web Intelligence, Mining and Semantics. pp.~1--8 (2018)

\bibitem{faizan2017multiple}
Faizan, A., Lohmann, S., Modi, V.: Multiple choice question generation for slides. In: Computer Science Conference for University of Bonn Students. pp.~1--6 (2017)

\bibitem{gangemi2006modelling}
Gangemi, A., Catenacci, C., Ciaramita, M., Lehmann, J.: Modelling ontology evaluation and validation. In: European semantic web conference. pp. 140--154. Springer (2006)

\bibitem{gavspar2023evaluation}
Ga{\v{s}}par, A., Grubi{\v{s}}i{\'c}, A., {\v{S}}ari{\'c}-Grgi{\'c}, I.: Evaluation of a rule-based approach to automatic factual question generation using syntactic and semantic analysis. Language resources and evaluation  \textbf{57}(4),  1431--1461 (2023)

\bibitem{hartmann2005d1}
Hartmann, J., Spyns, P., Giboin, A., Maynard, D., Cuel, R., Su{\'a}rez-Figueroa, M.C., Sure, Y.: D1. 2.3 methods for ontology evaluation. EU-IST Network of Excellence (NoE) IST-2004-507482 KWEB Deliverable D  \textbf{1} (2005)

\bibitem{hlomani2014approaches}
Hlomani, H., Stacey, D.: Approaches, methods, metrics, measures, and subjectivity in ontology evaluation: A survey. Semantic Web  \textbf{1}(5),  1--11 (2014)

\bibitem{keet2019african}
Keet, C.M.: The {A}frican wildlife ontology tutorial ontologies. Journal of Biomedical Semantics  \textbf{11}(4),  1--11 (2020)

\bibitem{kuo2003difficulty}
Kuo, R., Lien, W.P., Chang, M., Heh, J.S.: Difficulty analysis for learners in problem solving process based on the knowledge map. In: the 3rd IEEE International Conference on Advanced Technologies. pp. 386--387. IEEE (2003)

\bibitem{kuo2004analyzing}
Kuo, R., Lien, W.P., Chang, M., Heh, J.S.: Analyzing problem's difficulty based on neural networks and knowledge map. Journal of Educational Technology \& Society  \textbf{7}(2),  42--50 (2004)

\bibitem{kurdi2020generation}
Kurdi, G.R.: Generation and mining of medical, case-based multiple choice questions. Phd thesis, The University of Manchester (2020)

\bibitem{kusuma2020automatic}
Kusuma, S.F., Siahaan, D.O., Fatichah, C.: Automatic question generation in education domain based on ontology. In: the International Conference on Computer Engineering, Network, and Intelligent Multimedia (CENIM). pp. 251--256. IEEE (2020)

\bibitem{leo2019ontology}
Leo, J., Kurdi, G., Matentzoglu, N., Parsia, B., Sattler, U., Forge, S., Donato, G., Dowling, W.: {Ontology-based generation of medical, multi-term MCQs}. International Journal of Artificial Intelligence in Education  \textbf{29}(2),  145--188 (2019)

\bibitem{lourdusamy2018review}
Lourdusamy, R., John, A.: A review on metrics for ontology evaluation. In: the International Conference on Inventive Systems and Control (ICISC). pp. 1415--1421. IEEE (2018)

\bibitem{lozano2004ontometric}
Lozano-Tello, A., G{\'o}mez-P{\'e}rez, A.: Ontometric: A method to choose the appropriate ontology. Journal of Database Management (JDM)  \textbf{15}(2),  1--18 (2004)

\bibitem{Maedche2002}
Maedche, A., Zacharias, V.: Clustering ontology-based metadata in the semantic web. In: Elomaa, T., Mannila, H., Toivonen, H. (eds.) Principles of Data Mining and Knowledge Discovery. pp. 348--360. Springer Berlin Heidelberg, Berlin, Heidelberg (2002)

\bibitem{papasalouros2018semantic}
Papasalouros, A., Chatzigiannakou, M.: Semantic web and question generation: An overview of the state of the art. In: the International Association for Development of the Information Society ({IADIS}) International Conference on e-Learning. pp. 189--192. ERIC (2018)

\bibitem{papasalouros2008automatic}
Papasalouros, A., Kanaris, K., Kotis, K.: Automatic generation of multiple choice questions from domain ontologies. In: IADIS e-Learning 2008 conference. vol.~1, pp. 427--434 (2008)

\bibitem{park2011ontology}
Park, J., Oh, S., Ahn, J.: Ontology selection ranking model for knowledge reuse. Expert Systems with Applications  \textbf{38}(5),  5133--5144 (2011)

\bibitem{Pfleeger2005}
Pfleeger, S.L.: Software Engineering: Theory and Practice. Prentice Hall PTR, USA, 2nd edn. (2001)

\bibitem{pittet2015exploiting}
Pittet, P., Barth{\'e}l{\'e}my, J.: Exploiting users’ feedbacks-towards a task-based evaluation of application ontologies throughout their lifecycle. In: the International conference on knowledge engineering and ontology development. vol.~2, pp. 263--268. SCITEPRESS (2015)

\bibitem{raad2015survey}
Raad, J., Cruz, C.: A survey on ontology evaluation methods. In: International conference on knowledge engineering and ontology development. vol.~2, pp. 179--186. SciTePress (2015)

\bibitem{raboanary2022architecture}
Raboanary, T., Keet, C.M.: An architecture for generating questions, answers, and feedback from ontologies. In: the Research Conference on Metadata and Semantics Research. pp. 135--147. Springer (2022)

\bibitem{raboanary2021generating}
Raboanary, T., Wang, S., Keet, C.M.: Generating answerable questions from ontologies for educational exercises. In: the Research Conference on Metadata and Semantics Research. pp. 28--40. Springer (2021)

\bibitem{stasaski2017multiple}
Stasaski, K., Hearst, M.A.: Multiple choice question generation utilizing an ontology. In: the 12th Workshop on Innovative Use of NLP for Building Educational Applications. pp. 303--312 (2017)

\bibitem{tartir2005ontoQA}
Tartir, S., Arpinar, I.B., Moore, M., Sheth, A.P., Aleman-Meza, B.: {OntoQA}: Metric-based ontology quality analysis. In: Proceedings of IEEE Workshop on Knowledge Acquisition from Distributed, Autonomous, Semantically Heterogeneous Data and Knowledge Sources (2005)

\bibitem{tartir2010ontological}
Tartir, S., Arpinar, I.B., Sheth, A.P.: Ontological evaluation and validation. In: Theory and applications of ontology: Computer applications, pp. 115--130. Springer (2010)

\bibitem{e2016modeling}
Vinu, E.V., Alsubait, T., Kumar, P.S.: Modeling of item-difficulty for ontology-based {MCQs} (2016), \url{https://arxiv.org/abs/1607.00869}

\bibitem{vinu2015novel}
Vinu, E.V., Kumar, P.: {A novel approach to generate MCQs from domain ontology: Considering DL semantics and open-world assumption}. Journal of Web Semantics  \textbf{34},  40--54 (2015)

\bibitem{venu2020difficulty}
Vinu, E.V., Kumar, P.: Difficulty-level modeling of ontology-based factual questions. Semantic Web  \textbf{11}(6),  1023--1036 (2020)

\bibitem{ev2017automated}
Vinu, E.V., Kumar, P.: Automated generation of assessment tests from domain ontologies. Semantic Web  \textbf{8}(6),  1023--1047 (2017)

\bibitem{Wang2019OntologyST}
Wang, S.: Ontology specifications to generate questions (nd), \url{https://api.semanticscholar.org/CorpusID:247410247}

\bibitem{yu2009requirements}
Yu, J., Thom, J.A., Tam, A.: Requirements-oriented methodology for evaluating ontologies. Information Systems  \textbf{34}(8),  766--791 (2009)

\end{thebibliography}

\end{document}